\documentclass[11pt,letterpaper]{article}

\usepackage[T1]{fontenc}
\usepackage[utf8]{inputenc}
\usepackage{lmodern}
\usepackage[margin=1in]{geometry}
\usepackage{amsmath,amssymb}
\usepackage{graphicx}
\graphicspath{{figures/}}
\usepackage{subfig}
\usepackage{microtype}
\usepackage[authoryear,round]{natbib}
\usepackage[hidelinks]{hyperref}

\title{Self-attention summary networks for subsurface velocity-model building from common-image gathers}

\author{Shiqin Zeng, Yunlin Zeng, Abhinav Prakash Gahlot,\\
Zijun Deng, and Felix J. Herrmann\\[0.5em]
\small Georgia Institute of Technology}
\date{}

\begin{document}
\maketitle
\begin{abstract}
Common-image gathers (CIGs) contain physically meaningful information about velocity-model errors through reflector focusing and residual moveout, but in conventional imaging workflows they are typically used only as diagnostic tools. In this work, we propose a multiscale self-attention summary network that maps high-dimensional 3D CIG volumes into compact conditioning embeddings for probabilistic subsurface velocity inversion. These learned embeddings preserve offset-dependent kinematic structure and spatial coherence while reducing variability caused by background-velocity mismatch. Conditioned on these summary embeddings, a flow-matching model learns a transport from a Gaussian source distribution to the posterior distribution of plausible velocity fields. Numerical experiments show that, compared with direct conditioning on raw CIGs, the proposed summary network improves posterior velocity inference. In particular, the multiscale attention design provides greater robustness to background-model mismatch, yielding more accurate posterior reconstructions and lower predictive uncertainty.
\end{abstract}
\section{Introduction}
Velocity model building remains a central challenge in seismic imaging because the inverse problem is high-dimensional, nonlinear, and non-unique \citep{symes2008mva}. Conventional optimization based approaches such as full-waveform inversion (FWI) can recover high-resolution subsurface models when supported by adequate low-frequency content and a sufficiently accurate initial model, but they remain sensitive to initialization, prone to cycle skipping, and typically yield a single deterministic estimate \citep{virieux2009overview,symes2008mva}. In contrast, common-image gathers (CIGs) contain offset-dependent information about reflector focusing and residual moveout, making them informative indicators of velocity-model mismatch \citep{biondi2006cig,biondi2004adcig}.

Despite being constructed under imperfect background velocity-models, CIGs retain structured, mismatch-dependent patterns such as residual moveout and defocusing, which remain informative for velocity inference and are also exploited in uncertainty-aware inversion frameworks based on subsurface extensions \citep{yin2024wise}. Building on this perspective, we use CIGs as conditioning variables for probabilistic velocity-model inversion by combining a self-attention summary encoder for high-dimensional CIG representations with a conditional flow-matching model for posterior velocity-model generation \citep{lipman2023flow}. The resulting framework learns a distribution over plausible velocity fields conditioned on the compact latent representation of the CIG, improving both reconstruction accuracy and uncertainty quantification.  

\section{Method}
\subsection{3D CIG representation}
We use \emph{subsurface-offset} common-image gathers (CIGs), shown in the middle column of Fig.~\ref{fig:summary_output}, as the extended-image representation for migration-velocity analysis \citep{biondi2004adcig,biondi2006cig}. Let the observed seismic data be denoted by $\mathbf{d}_{\mathrm{obs}}$, generated from the true subsurface model $\mathbf{m}_{\mathrm{gt}}$ through the forward operator $\mathcal{F}$, i.e., $\mathbf{d}_{\mathrm{obs}}=\mathcal{F}(\mathbf{m}_{\mathrm{gt}})$. Given a background model $\mathbf{m}_0$, we define the time-domain extended imaging condition as in Eq.~\ref{eq:cig}
\begin{equation}
I(\mathbf{x},\mathbf{h};\mathbf{m}_0)
=
\sum_{s}\int
u_s(\mathbf{x}-\mathbf{h},t;\mathbf{m}_0)\,
u_r(\mathbf{x}+\mathbf{h},t;\mathbf{d}^{(s)}_{\mathrm{obs}},\mathbf{m}_0)\,dt,
\label{eq:cig}
\end{equation}
where $\mathbf{x}$ is the image location, $\mathbf{h}$ is the subsurface half-offset, and $s$ indexes the source experiment. Here, $u_s$ denotes the source wavefield propagated in $\mathbf{m}_0$, and $u_r$ denotes the receiver wavefield obtained by back-propagating the observed data in the same background model. When $\mathbf{m}_0$ is kinematically consistent with $\mathbf{m}_{\mathrm{gt}}$, reflected energy focuses near zero offset ($\mathbf{h}=\mathbf{0}$); otherwise, model error produces residual moveout and defocusing across offsets. In practice, $I(\mathbf{x},\mathbf{h};\mathbf{m}_0)$ is discretized as a real-valued tensor $\mathbf{y}\in\mathbb{R}^{D\times H\times W}$, where $D$ is the number of sampled offsets and $(H,W)$ are the spatial dimensions. Each spatial location is therefore associated with an offset-dependent response vector that encodes local kinematic and structural information relative to the background model. This representation, however, remains highly redundant across both offset and space.
\begin{figure}[htbp]
    \centering
    \includegraphics[width=0.85\textwidth]{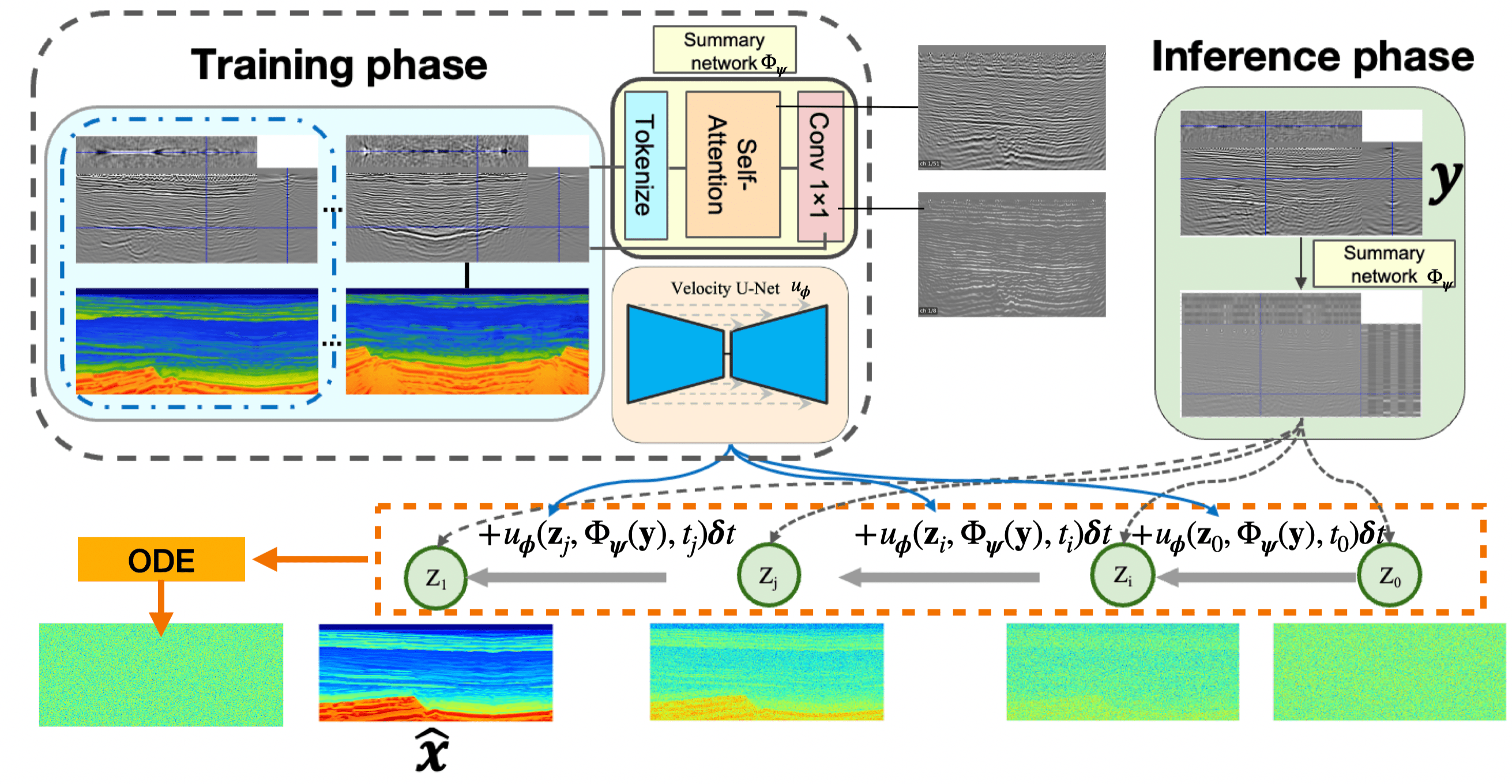}
    \caption{Overall workflow of the proposed method.}
    \label{fig:workflow}
\end{figure}

\subsection{Multiscale self-attention summary network}
To obtain a compact and robust seismic representation of CIGs, we learn a summary encoder $\Phi_\psi(\cdot)$ that maps the input CIGs to a low-dimensional latent embedding. We flatten the spatial grid into tokens indexed by $i \in \{1,\dots,HW\}$, where each index corresponds to a spatial location and is associated with an offset-response vector from the raw CIGs. Specifically, let $\mathbf{y}_i \in \mathbb{R}^{D}$ denote the offset vector at token $i$. We then apply local windowed self-attention at multiple window sizes, since the query, key, and value vectors are all computed from the same token set \citep{vaswani2017attention}. For a token index $i$ and a neighboring token index $j$ within a local window $\mathcal{W}(i)$, we define $\mathbf{q}_i = W_Q \mathbf{y}_i$, $\mathbf{k}_j = W_K \mathbf{y}_j$, and $\mathbf{v}_j = W_V \mathbf{y}_j$, where $W_Q$, $W_K$, and $W_V$ are learnable projection matrices. The attention weights from token $i$ to token $j$ are given by
\begin{equation}
\alpha_{ij}
=
\frac{
\exp\!\left( \mathbf{q}_i^\top \mathbf{k}_j / \sqrt{d_a} \right)
}{
\sum_{j' \in \mathcal{W}(i)}
\exp\!\left( \mathbf{q}_i^\top \mathbf{k}_{j'} / \sqrt{d_a} \right)
},
\label{eq:attention-weight}
\end{equation}
where $\mathcal{W}(i)$ denotes the local spatial window centered at token $i$, and $d_a$ is the attention dimension. The attended feature at token $i$ is then $\tilde{\mathbf{y}}_i = \sum_{j \in \mathcal{W}(i)} \alpha_{ij}\,\mathbf{v}_j$. This mechanism allows each spatial location to aggregate information from nearby locations with similar offset-dependent responses, reinforcing spatially coherent reflector structure while suppressing locally inconsistent artifacts. To encode structure across different spatial extents, we employ multiple attention window sizes, similar in spirit to window-based self-attention architectures\citep{liu2021swin}. This allows the model to capture both local kinematic variations and larger-scale continuity. After attention-based aggregation, the features are compressed through a learned channel-mixing projection to reduce redundancy while preserving physically relevant information. The final embedding, illustrated by the summary-network outputs in Fig.~\ref{fig:summary_output}, serves as a compact, mismatch-aware conditioning signal for downstream probabilistic velocity generation.

\subsection{Velocity-model generation via conditional flow matching}
Given the summary encoder, we model the conditional distribution of subsurface velocity models using conditional flow matching. We jointly learn the encoder $\Phi_{\psi}$ and the conditional vector field $u_{\phi}$ by minimizing Eq.~\eqref{eq:cfm}:
\begin{equation}
\mathcal{L}(\psi,\phi)
=
\mathbb{E}_{t,\,\mathbf{x}_t,\,\mathbf d^{\mathrm{CIG}}_{\mathrm{seis}}}
\left[
\left\|
u_{\phi}\!\left(\mathbf{x}_t,t,\Phi_{\psi}\!\left(\mathbf y\right)\right)
-
u_t\!\left(\mathbf{x}_t \mid \Phi_{\psi}\!\left(\mathbf y\right)\right)
\right\|_2^2
\right]
\label{eq:cfm}
\end{equation}

In this work, we use the rectified-flow probability path \citep{liu2022rectified,lipman2023flow}, defined by the linear interpolation $\mathbf{x}_t = (1-t)\mathbf{x}_0 + t \mathbf{x}_1,$ where $\mathbf{x}_0 \sim \mathcal{N}(0,I)$ and $\mathbf{x}_1 \sim p_{\mathrm{\mathbf{m}}}$ of subsurface velocity modeling. The corresponding target vector field is $u_t = \frac{d\mathbf{x}_t}{dt} = \mathbf{x}_1 - \mathbf{x}_0$. While we adopt the linear rectified-flow path in this work, other choices of probability paths or optimal transport constructions can also be used. At inference, we sample $\mathbf{x}_0 \sim \mathcal{N}(0,I)$ and generate the subsurface velocity model by integrating the learned conditional ODE in Eq.~\eqref{eq:ode_sample}:
\begin{equation}
\mathbf{x}_1
=
\mathbf{x}_0
+
\int_0^1
u_{\phi}\!\left(\mathbf{x}_t,t,\Phi_{\psi}\!\left(\mathbf y\right)\right)\,dt.
\label{eq:ode_sample}
\end{equation}

Accordingly, ODE-based sampling provides a practical and theoretically justified choice: it yields a deterministic realization of the learned transport while preserving the probabilistic interpretation via random Gaussian initialization. Since the objective is to match the conditional marginal distribution, the probability-flow ODE is sufficient provided that it transports the source distribution to the same target marginals \citep{song2021score,albergo2025stochasticinterpolants}. Compared with explicitly simulating an SDE, this results in a simpler and often more computationally efficient inference procedure, while still enabling uncertainty quantification through repeated sampling. An overview of the full pipeline is shown in Fig.~\ref{fig:workflow}.
\section{Experiments and results}
\subsection{Compact latent representation from summary encoding}
\begin{figure}[htbp]
    \centering
    \subfloat[Singular value spectra across compressed channel sizes.\label{fig:singular_comparison}]{
        \includegraphics[width=0.48\textwidth]{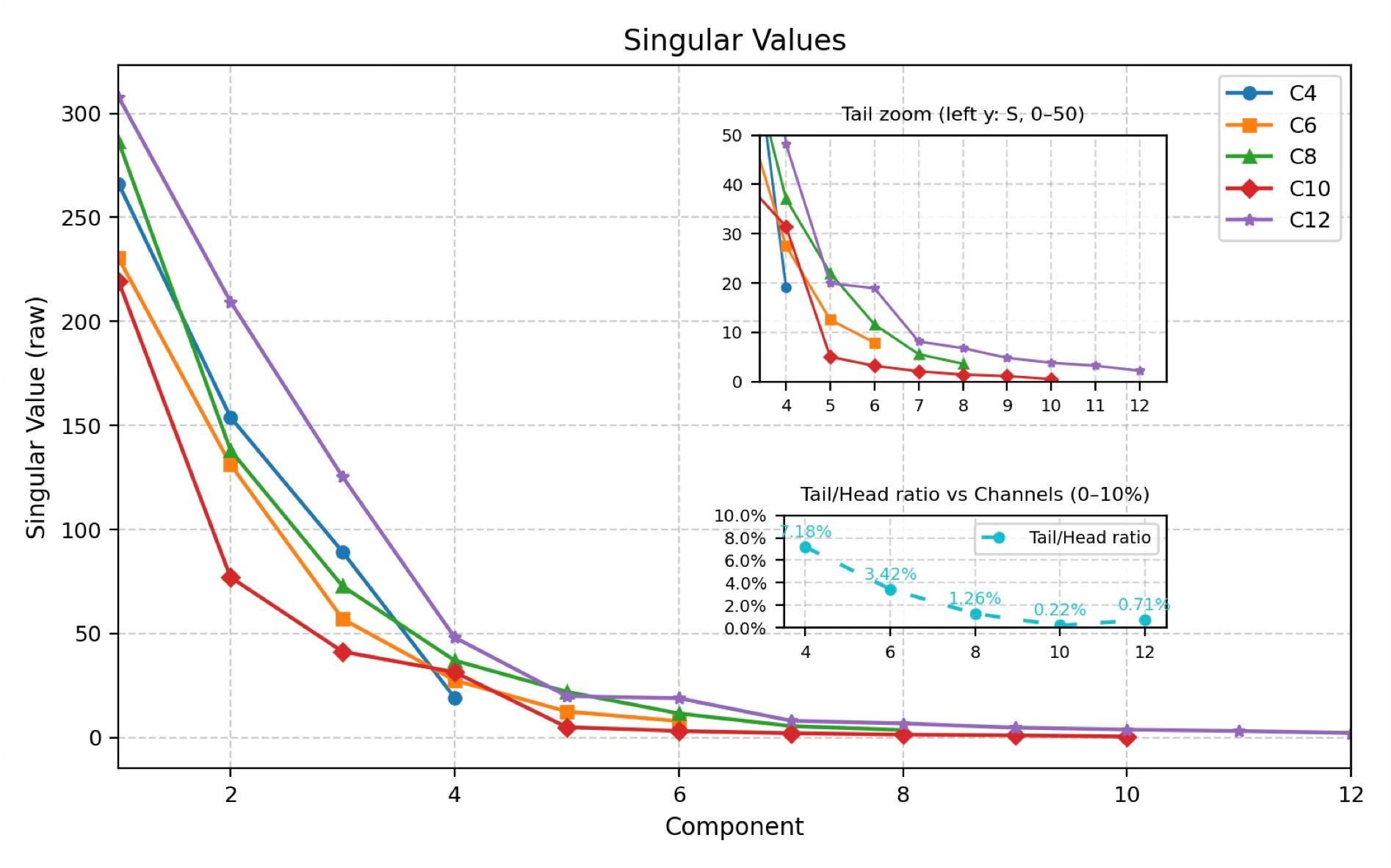}
    }
    \hfill
    \subfloat[Downstream performance comparison across channel sizes.\label{fig:model_performance_comparison}]{
        \includegraphics[width=0.48\textwidth]{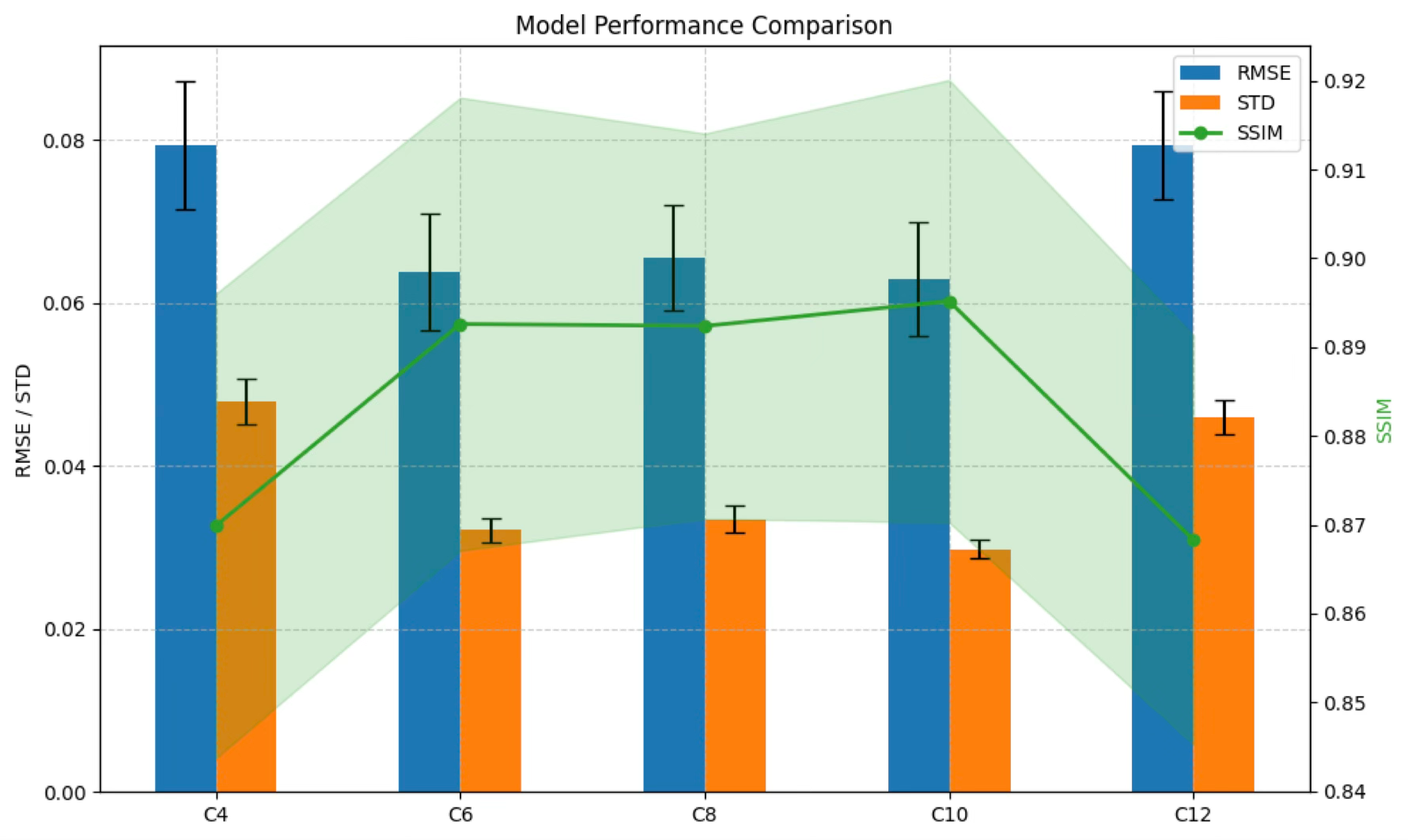}
    }
    \caption{Selection of compressed channel size based on spectral compactness and downstream predictive performance. }
    \label{fig:compressed_channel_selection}
\end{figure}

We first examine how the compressed channel size $C\ll D$, affects latent compactness and downstream predictive performance. As shown in Fig.~\ref{fig:compressed_channel_selection}, increasing $C$ from 4 to 10 leads to a monotonic decrease in the singular value ratio $(\sigma_{\min}/\sigma_{\max})$, from $7.2\%$ to $0.22\%$. We use this ratio as a spectral proxy for latent compactness because smaller values indicate stronger suppression of trailing singular directions and hence a more concentrated spectrum with lower effective dimensionality \citep{roy2007effective,jing2020implicit}. Across $C=4$ to $10$, downstream performance remains stable overall and improves relative to $C=4$, with lower RMSE, reduced predictive standard deviation, and higher SSIM for $C=6$, $8$, and $10$. These results suggest that $C=6$--$10$ provides a favorable balance between latent compactness and predictive quality. In the remainder of this work, we use $C=8$ as the default latent dimensionality. By contrast, increasing to $C=12$ yields a larger singular value ratio and degraded downstream performance, indicating diminishing returns beyond the preferred range.

\subsection{Behavior under background-velocity model errors}
\begin{figure}[htbp]
    \centering
    \includegraphics[width=1.0\textwidth]{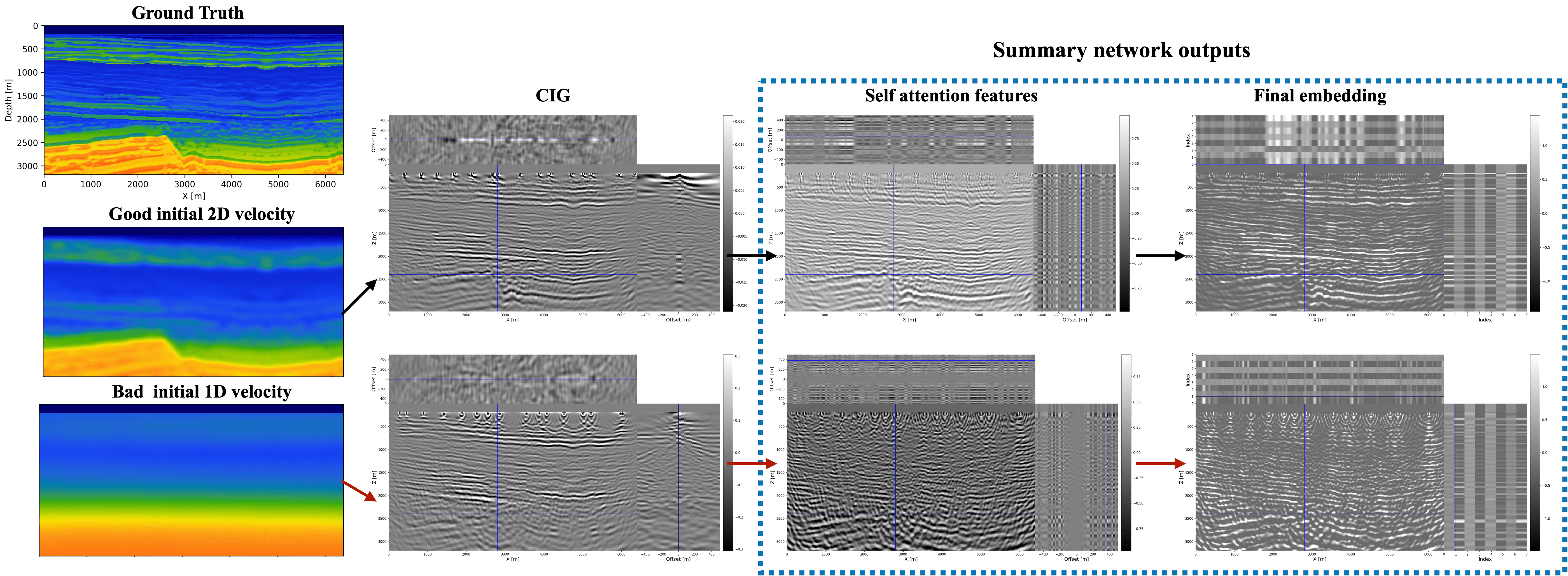}
    \caption{Summary-network response under different background models. }
    \label{fig:summary_output}
\end{figure}

The summary network behaves differently under different initial background-velocity models. As shown in Fig.~\ref{fig:summary_output}, with the 2D smoothed initial background, the input CIGs remain relatively well focused near zero offset, and the attention features preserve broad offset-dependent structure. With the 1D initial background, the input CIGs show stronger defocusing and noisier responses around zero offset. In this case, the attention features shift their emphasis away from near-zero-offset-dominated responses and redistribute it across other offset-related components.

These observations suggest that the summary network does not directly recover the correct focusing, but instead learns a mismatch-aware redistribution of offset information. Self-attention adaptively aggregates information from neighboring spatial locations according to the input feature patterns, while reweighting and mixing offset-dependent responses within each spatial location. As a result, the summary network adjusts its representation to the mismatch reflected in the input CIGs and produces a summary embedding that is more robust for downstream probabilistic velocity generation.

\subsection{Effect of multiscale attention windows}
\begin{figure}[htbp]
    \centering
    \includegraphics[width=\linewidth]{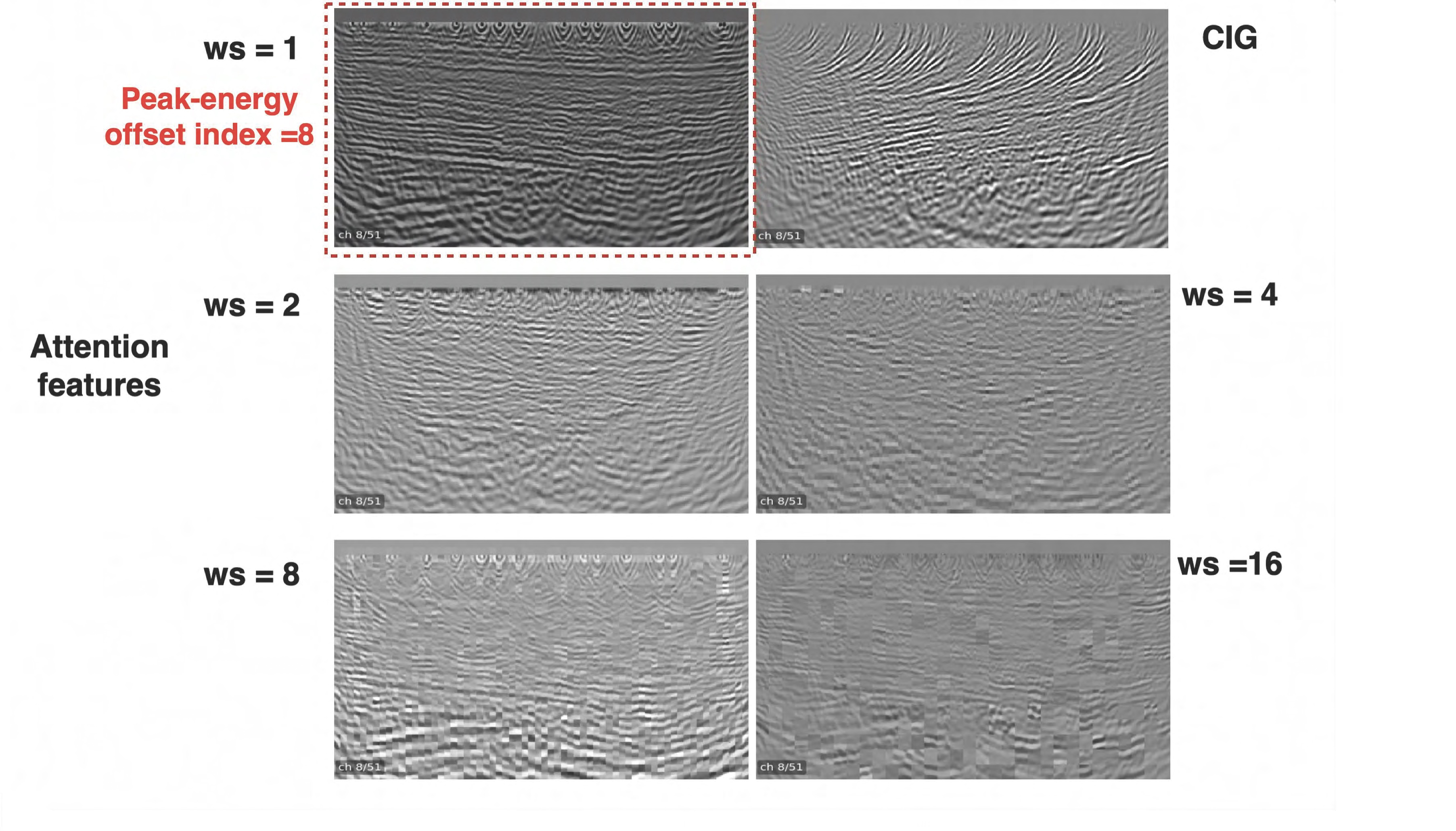}
    \caption{Attention features at the selected peak-energy offset index ($=8$) for different attention window sizes.}
    \label{fig:ws1}
\end{figure}

Fig.~\ref{fig:ws1} illustrates how the attention window size controls the scale of information retained in the summary features under the 1D initial background model. At $ws=1$, the feature largely preserves high-frequency local detail, since no spatial neighborhood aggregation is introduced. At $ws=2$, limited neighborhood mixing begins to suppress local fluctuations while retaining fine-scale structure. At $ws=4$, the feature becomes smoother and more coherent while still preserving consistent reflector geometry. As the window size increases further, the representation becomes progressively coarser. At $ws=8$, the feature emphasizes broader structural continuity, whereas at $ws=16$ it becomes overly smooth and blocky, with less local detail.

Overall, these feature patterns indicate that different window sizes capture complementary information. Smaller windows preserve sharper local responses, while larger windows provide broader structural context. This motivates the use of multiscale attention to combine fine-scale detail with larger-scale contextual aggregation.

\begin{figure}[htbp]
    \centering \includegraphics[width=0.96\textwidth]{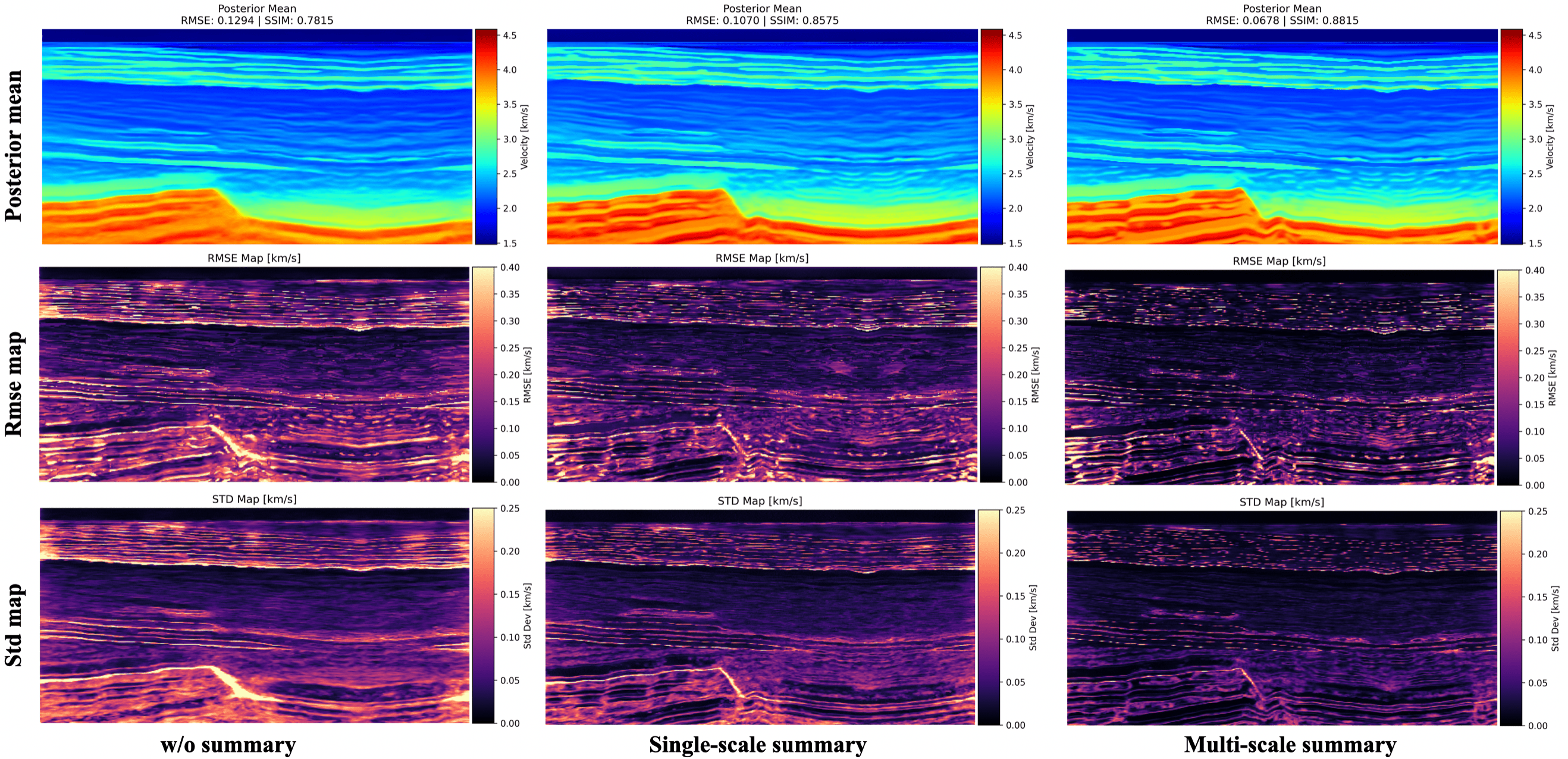}
    \caption{Comparison of posterior velocity inference under three conditioning strategies.} \label{fig:summary_compare}
\end{figure}
\subsection{Summary conditioning for posterior inference}
We next evaluate how the quality of the conditioning representation affects posterior velocity-model inference from CIGs. Fig.~\ref{fig:summary_compare} compares three configurations under the same 1D initial background model: using the raw CIG directly as conditioning, using a single-scale summary network, and using a multiscale summary network. Moving from raw CIG conditioning to single-scale summary conditioning to multiscale summary conditioning, the posterior mean becomes progressively closer to the ground-truth velocity model, with a monotonic reduction in RMSE and a corresponding increase in SSIM.
The spatial RMSE maps further show that, when the raw CIG is used directly as conditioning, reconstruction errors are more broadly distributed and more pronounced near layered interfaces and structurally complex regions. Introducing a summary network makes these errors more localized, while the multiscale summary network yields the cleanest error profile overall. Similarly, the posterior standard-deviation maps indicate that uncertainty is progressively reduced from the raw-CIG conditioning baseline to the single-scale and multiscale summary variants. The multiscale summary representation yields the most concentrated and best-structured uncertainty field, indicating that combining multiple attention scales improves both posterior accuracy and uncertainty control. Together, Fig.~\ref{fig:ws1} and Fig.~\ref{fig:summary_compare} show that multiscale attention improves the conditioning representation at the feature level and translates this improvement into more accurate and consistent posterior velocity estimates.
\section{Conclusion and future work}
We propose a probabilistic framework for subsurface velocity-model building that combines a multiscale self-attention summary network that acts on subsurface-offset Common Image Gathers (CIGs) with conditional flow matching. The summary network converts high-dimensional CIGs into compact conditioning representations, and the flow-matching model uses them to generate posterior velocity models. The results show that this combination improves robustness under background-model errors and yields more accurate posterior samples and more concentrated uncertainty fields than direct raw CIG conditioning. Overall, multiscale self-attention and conditional flow matching provide an effective approach for uncertainty-aware velocity inference from CIGs. Future work will test generalization across diverse initial background-velocity models by training on CIGs with different velocity errors and evaluating on unseen backgrounds. This would provide stronger evidence that the summary network learns adaptive mismatch-aware representations.
\section*{Acknowledgments}
This research was carried out with the support of the Georgia Research Alliance and the partners of the ML4Seismic Center.

\bibliographystyle{plainnat}
\bibliography{references}

\end{document}